\documentclass[runningheads]{llncs} 
\usepackage[T1]{fontenc} 
\usepackage{graphicx} 
\usepackage{float}

\begin{document}
\title{ Intelligent Spark Agents: A Modular LangGraph Framework for Scalable, Visualized, and Enhanced Big Data Machine Learning Workflows }

\author{Jialin Wang\inst{1}  \and
Zhihua Duan\inst{2 }   }
 
\institute{
Executive Vice President,Ferret Relationship Intelligence\\Burlingame, CA 94010, USA \\
\email{jialinwangspace@gmail.com}\\
\url{https://www.linkedin.com/in/starspacenlp/} 
\and
Intelligent Cloud Network Monitoring Department \\
China Telecom Shanghai Company,Shanghai, China\\
\email{duanzh.sh@chinatelecom.cn}\\
}

\maketitle              
\begin{abstract}
This paper presents a Spark-based modular LangGraph framework, designed to enhance machine learning workflows through scalability, visualization, and intelligent process optimization. At its core, the framework introduces Agent AI, a pivotal innovation that leverages Spark’s distributed computing capabilities and integrates with LangGraph for workflow orchestration.

Agent AI facilitates the automation of data preprocessing, feature engineering, and model evaluation while dynamically interacting with data through Spark SQL and DataFrame agents. Through LangGraph’s graph-structured workflows, the agents execute complex tasks, adapt to new inputs, and provide real-time feedback, ensuring seamless decision-making and execution in distributed environments. This system simplifies machine learning processes by allowing users to visually design workflows, which are then converted into Spark-compatible code for high-performance execution.

The framework also incorporates large language models through the LangChain ecosystem, enhancing interaction with unstructured data and enabling advanced data analysis. Experimental evaluations demonstrate significant improvements in process efficiency and scalability, as well as accurate data-driven decision-making in diverse application scenarios.

This paper emphasizes the integration of Spark with intelligent agents and graph-based workflows to redefine the development and execution of machine learning tasks in big data environments, paving the way for scalable and user-friendly AI solutions. 

\end{abstract}
 
\section{Introduction}
The development of information technology brings convenience to life and fast-growing data. With the maturity of big data analysis technology represented by machine learning, big data has tremendous effect on social and economic life and provided a lot of help for business decision-making. For example, in the e-commerce industry, Taobao recommends suitable goods professionally to users after analyzing from large amounts of transaction data; in the advertising industry, online advertising predicts users' preferences by tracking users' clicks to improve users' experience.

However, the traditional business relational data management system has been unable to deal with the characteristics of big data including large capacity, diversity, and high-dimension . \cite{1} In order to solve the problem of big data analysis, distributed computing has been widely used, among which the Apache Hadoop \cite{2} is one of the widely used distributed systems in recent years. Hadoop adopts MapReduce as a rigorous computing framework. The emergence of Hadoop has promoted the popularity of large-scale data processing platforms. Spark \cite{3}, a big data architecture developed by AMPLab of the University of Berkeley, is also widely used. Spark integrates batch analysis, flow analysis, SQL processing, graph analysis, and machine learning applications. Compared with Hadoop, Spark is fast, flexible, and fault-tolerant, which is the ideal choice to run machine learning analysis programs. However, Spark is a tool for developers, which requires analysts to have certain computer skills and spend a lot of time creating, deploying  and maintaining systems.

The results of machine learning depend heavily on data quality and model logic, so this paper designs and implements a Park-based flow machine learning analysis tool in order to enable analysts to concentrate on the process itself and not spend energy on compiling, running, and parallelizing the analysis program. Formally, each machine learning analysis task is decomposed into different stages and is composed of components, which reduces the user's learning cost. In technology, general algorithms are encapsulated into component packages for reuse, and the training process is differentiated by setting parameters, which reduces the time cost of creating machine learning analysis programs. Users can flexibly organize their own analysis process by dragging and pulling algorithm components to improve the efficiency of application creation and execution.

Spark, as a powerful distributed computing system, has significant advantages in the field of big data processing and analysis. With the rapid development of large model technology, the application of Spark combined with large model agents has gradually become a hot topic in research. Large model agents can more accurately perceive the environment, react and make judgments, and form and execute decisions. Supported by Spark, large model agents can process large-scale datasets, achieving efficient data analysis and decision-making. LangGraph is an agent workflow framework officially launched by LangChain, which defines workflows based on graph structures, supports complex loops and conditional branches, and provides fine-grained agent control. The integration of Spark large model agents with the advanced tools and workflow management of the Langchain and LangGraph frameworks is driving innovative applications in various fields, including large model data analysis.

This paper will show the characteristics of the tool by comparing with the currently existing products, and then make a detailed description on the system architecture design, business model with use cases, and function operation by in-depth system modules. At the same time, this paper will also provide a technical summary and look forward to the future research directions based on LangGraph for Spark agents.

\section{Brief Introduction of Related Technologies}

\subsection{AML Machine Learning Service} 
Azure Machine Learning (AML) \cite{4} is a Web-based machine learning service launched by Microsoft on its public cloud Azure, which contains more than 20 algorithms for classification, regression, and clustering based on supervised learning and unsupervised learning, and is still increasing. However, AML is based on Hadoop and can only be used in Azure. Different from AML, the tool in this paper is designed and implemented based on Spark, therefore, it can be deployed on different virtual machines or cloud environments.
 
\subsection{Responsive Apache Zeppline Based on Spark}

Apache Zeppline \cite{5} is a Spark-based responsive data analysis system, whose goal is to build interactive, visualized, and shareable Web applications that integrate multiple algorithmic libraries. It has become an open source, notebook-based analysis tool that supports a large number of algorithmic libraries and languages. However, Zeppelin does not provide a user-friendly graphical interface, and all analyzers require users to write scripts to submit and run, which improves the user's programming technical requirements. The tools mentioned in this paper provide component graphics tools and a large number of machine learning algorithms, which make users simply and quickly define the machine learning process and run to get results.

\subsection{Haflow big data Analysis Service Platform}
In reference \cite{6}, Haflow, a big data analysis service platform, is introduced. The system uses component design, which can make users drag and drop the process analysis program, and open an extended interface, which enables developers to create custom analysis algorithm components. At present, Haflow only supports MapReduce algorithm components of Hadoop platform but the tool mentioned in this paper is based on Haflow, so that it can support Spark's component application and provides a large number of machine learning algorithms running in Spark environment.

\subsection{ Large Language  Models Usher in a New Era of Data Management and Analysis}
With the rapid advancement of artificial intelligence technology, large model technology has become a hot topic in the field of AI today. In the realm of data management and analysis, the emergence of large language models (LLMs), such as GPT-4o, Llama 3.2, ERNIE -4, GLM-4, and other large models, has initiated a new era filled with challenges. These large models possess powerful semantic understanding and efficient dialogue management capabilities, which have a profound impact on data ingestion, data lakes, and data warehouses. The natural language understanding functionality based on large language models simplifies data management tasks and enables advanced analysis of big data, promoting innovation and efficiency in the field of big data.

\subsection{LangChain and LangGraph Frameworks} 
LangChain is a large-scale model application development framework designed for creating applications based on Large Language Models (LLMs). It is compatible with a variety of language models and integrates seamlessly with OpenAI ChatGPT. The LangChain framework offers tools and agents that "chain" different components together to create more advanced LLM use cases, including Prompt templates, LLMs, Agents, and Memory, which help users to improve efficiency when facing challenges and generating meaningful responses.

LangGraph is a framework within the LangChain ecosystem that allows developers to define cyclic edges and nodes within a graph structure and provides state management capabilities. Each node in the graph makes decisions and performs actions based on the current state, enabling developers to design agents with complex control flows, including single agents, multi-agents, hierarchical structures, and sequential control flows, which can robustly handle complex scenarios in the real world.

The combination of Spark with LangChain and LangGraph allows data enthusiasts from various backgrounds to effectively engage in data-driven tasks. Through Spark SQL agents and Spark DataFrame agents, users are enabled to interact with, explore, and analyze data using natural language.

\section{Process-oriented Machine Learning Analysis Tool Based on Spark}

\subsection{Machine Learning Process}
This paper aims at designing a process machine learning tool for data analysts, so it is necessary to implement the functions of common machine learning processes. Machine learning can be divided into supervised learning and unsupervised learning, mainly depending on whether there are specific labels. Labels are the purpose of observation data or the prediction object. Observation data are the samples used to train and test machine learning models. Feature is the attribute of observation data. Machine learning algorithm mainly trains the prediction rules from the characteristics of observation data.

In practice, machine learning process is composed of a series of stages, including data preprocessing, feature processing, model fitting, and result verification or prediction. For example, classify a group of text documents including word segmentation, cleaning, feature extraction, training classification model, and output classification results \cite{7}.
\begin{figure}
\centering
\includegraphics[width=0.6\textwidth]{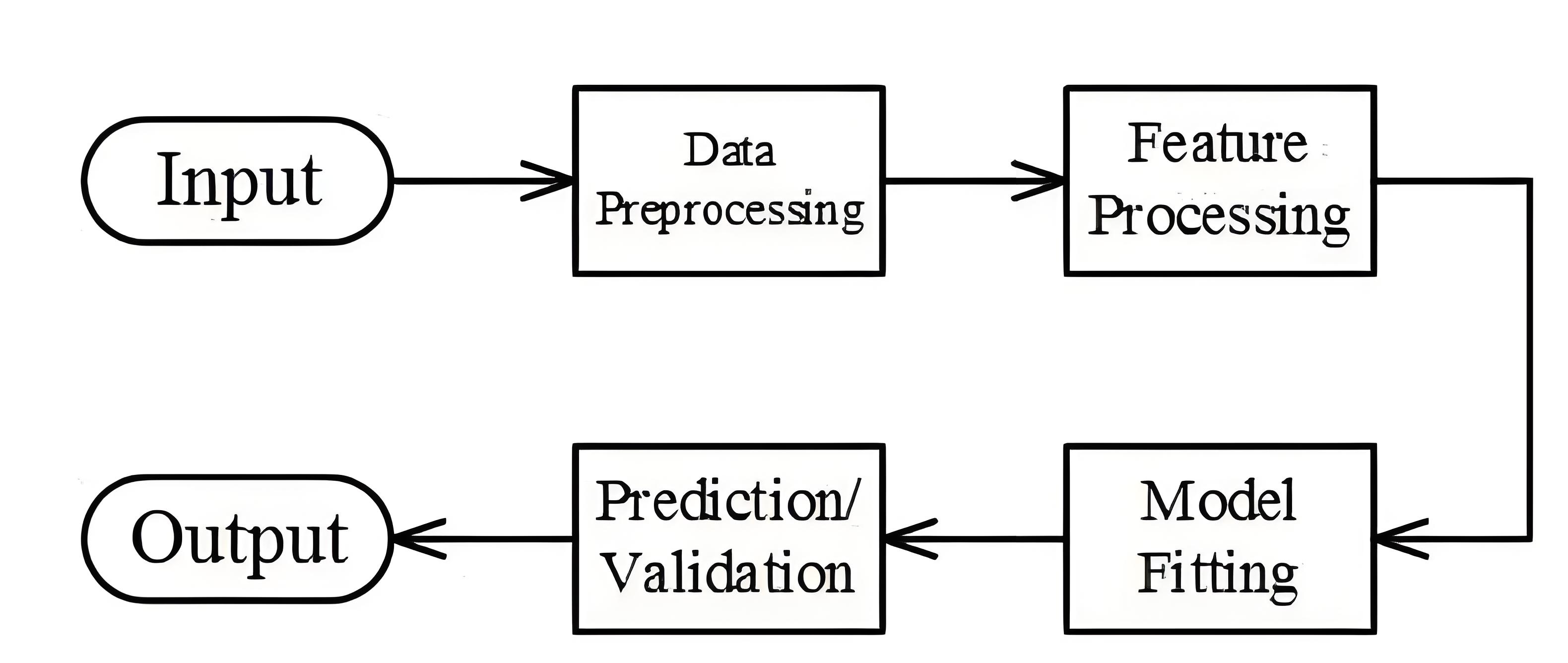}
\caption{Typical machine learning process.} \label{fig1}
\end{figure}

These stages can be seen as black-box processes and packaged into components. Although there are many algorithmic libraries or software that provide programs for each stage, these programs are seldom prepared for large-scale data sets or distributed environments, and these programs are not naturally supporting process-oriented, requiring developers to connect each stage to form a complete process. 

Therefore, while providing a large number of machine learning algorithm components, the system also needs to complete the function of automatic process execution, taking into account the operational efficiency of the process.

\subsection{Design Process of Business Module}
The system provides components to users as main business functions. The analysts can freely combine the existing components into different analysis processes. In order to be able to cover the commonly used machine learning processes, this system provides the following categories of business modules: input/output module, data preprocessing module, characteristic processing module, model fitting module, and the results predicted module. Different from other systems, the business module designed by this tool is using each stage of the process as a definition.

1. Input and output module. Used to realize data acquisition and writing, dealing with heterogeneous data sources, this module is the starting point and endpoint of the machine learning process. In order to be able to handle different types of data, this system provides data input or output functions of structured data (such as CSV data), unstructured data (such as TXT data), and semi-structured data (such as HTML).

2. Data preprocessing module. This module includes data cleaning, filtering, the join/fork, and type change, etc. Data quality determines the upper limit of the accuracy of the machine learning model, so it is necessary to improve the data preprocessing process before feature extraction. This module can clean up null or abnormal values, change data types, and filter out unqualified data.

3. Feature processing module. Feature processing is the most important link before modeling data, including feature selection and feature extraction. The system currently contains 25 commonly used feature processing algorithms.

\subsection{Feature Extraction and Classification}
Feature selection is a multi-dimensional feature selection. The most valuable feature is selected by the algorithm. The selected feature is a subset of the original feature. According to the selected algorithm, it can be divided into information gain selector, chi-square information selector, and Gini coefficient selector. 

Feature extraction is to transform the features of observed data into new variables according to a certain algorithm. Compared with data preprocessing, the rules of data processing are more complex. The extracted features are the mapping of the original features, including the following categories:

1. Standardized component. Standardization is an algorithm that maps numerical features of data to a unified dimension. Standardized features are unified to the same reference frame, which makes the training model more accurate and converges faster in the training process. Different standardized components use different statistics to map, such as normalizer components, Standard Scaler components, MinMax Scaler groups  and so on.

2. Text processing components. Text type features need to be mapped to new numeric type variables because they cannot be calculated directly. Common algorithms include TF-IDF components that index text by word segmentation, Tokenizer components for word segmentation, OneHotEncoder components for hot encoding, etc.

3. Dimension-reducing components. This kind of components compress the original feature information through a certain algorithm and express it with fewer features, such as PCA components of principal component analysis.

4. Custom UDF components. Users can input the function of SQL custom feature processing.

5. Model fitting module. Model training uses certain algorithms to learn data, and the obtained model can be used for subsequent data prediction. At present, the system provides a large number of supervised learning model components, which can be divided into classification models and regression models according to the different nature of observation data labels.

6. Results prediction module. This module includes two functions: results prediction and verification. 

Through the above general business modules, users can create a variety of common machine learning analysis processes in the system environment.

\subsection{System Architecture Design}
The system provides a user interface through Web, and the overall architecture is mainly based on the MVC framework. At the same time, it provides business modules of machine learning and execution modules of processes. The system architecture is shown in Figure 2.
\begin{figure}
\centering
\includegraphics[width=1.0\textwidth]{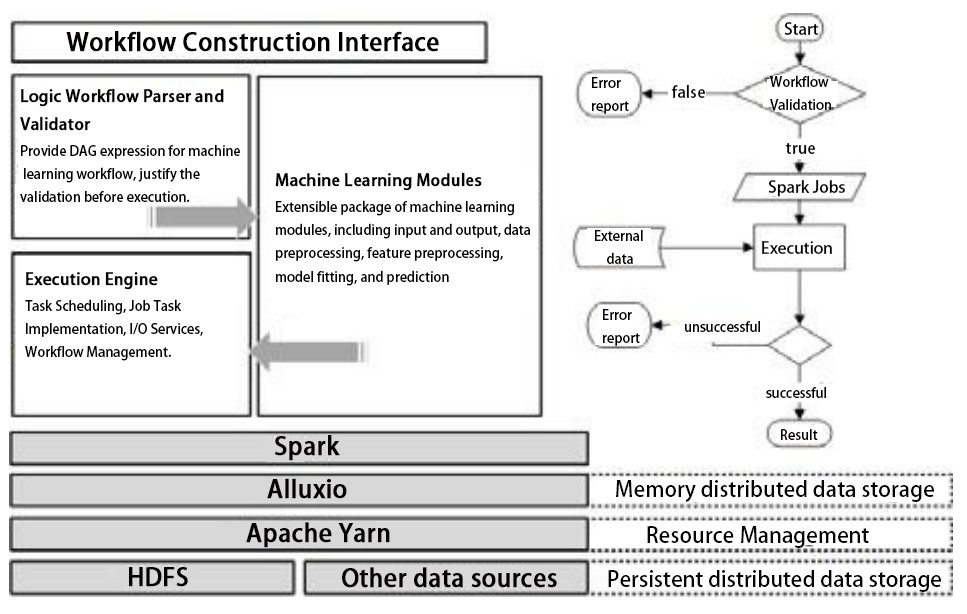}
\caption{System architecture diagram and flow chart.} \label{fig2}
\end{figure}

Users create formal machine learning processes through the Web interface provided by the system and submit them to the system. The system converts the received original processes into logical flow charts and verifies the validity of the flow charts. Validation of process validity is a necessary part of the analysis process before actual execution. When the process has obvious errors such as logic or data mismatch, it can immediately return the error, rather than wait for the execution of the corresponding component to report the error, which improves the efficiency of the system. 

The execution engine of the system is the key module, which implements the multi-user and multi-task process execution function. It translates the validated logical flow chart into the corresponding execution model, which is the data structure identifiable by the system and used to schedule the corresponding business components. The translation of the execution model is a complex process, which will be introduced in detail in Section 4.3.

\subsection{ Architecture Design of Spark Agent Based on LangGraph}
As shown in Figure 3, the architecture of the agent implementation that combines Apache Spark with LangChain and LangGraph is designed to enhance the level of intelligence in data processing and decision-making. This architectural diagram displays a Spark-based agent system capable of accomplishing complex tasks by integrating various technological components.

\begin{figure}
\centering
\includegraphics[width=1.0\textwidth]{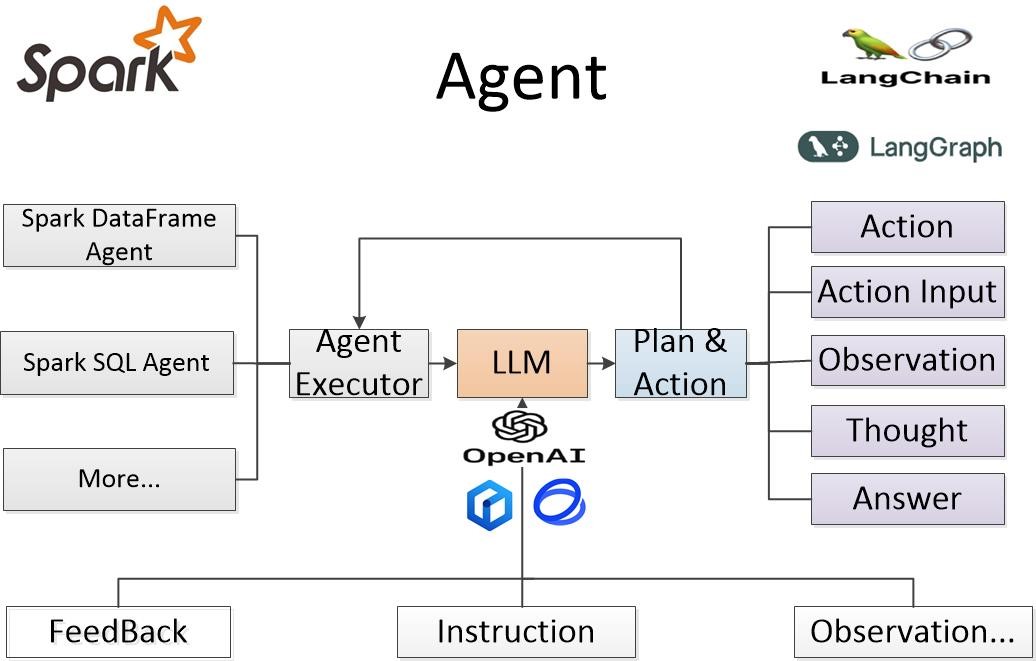}
\caption{ Architectural Design of Spark Agent Based on LangGraph.} \label{fig3}
\end{figure} 

LangChain and LangGraph frameworks are introduced on top of Spark. LangChain is a framework for building and deploying large language models, enabling agents to interact with users or other systems through natural language. LangGraph provides a graph-structured workflow, allowing agents to plan and execute tasks in a more flexible manner. The core of the agent is the LLM (Large Language Model), which is responsible for understanding and generating natural language. The LLM receives instructions (Instruction), observations (Observation), and feedback (FeedBack) from LangChain and integrates these with data insights from Spark to form thoughts (Thought). These thoughts are then translated into specific action plans (Plan and Action), which are executed by the action executor (Action Executor). 

The action executor is tightly integrated with Spark, processing structured data through Spark DataFrame Agent and Spark SQL Agent, performing complex data analysis and data processing tasks, and feeding the results back to the LLM. The LLM adjusts its action plans based on this feedback and new observations, forming a closed-loop learning and optimization process that further refines its performance and decision-making quality.

\section{Research on System Implementation and Key Technologies}
\subsection{Storage of Intermediate Data}

\subsubsection{Frame-based Contingent Storage Architecture}
In the whole process of machine learning, data is in the state of flow, and components with sequence dependence need to transfer intermediate data. In order to avoid the problem of heterogeneity of intermediate data, the system stipulates that components communicate with each other using a unified determinant storage structure based on DataFrame \cite{8}, which is a Spark-supported distributed data set and is classified as the main data set. Conceptually, it is similar to the "table" of relational database, but a lot of optimization has been done on its operation execution at Spark. In this way, the relationship of structured data is retained, special data attributes are defined, features and label are specified as the head of data required in the model fitting stage, so as to facilitate the validation and execution of the process. 

This determinant storage structure can be quickly persisted to the intermediate data storage layer by the whole system, and quickly restored to the required data objects when the later components are used.

\subsubsection{Alluxio Virtual Distributed Storage System}

Intermediate data needs different management in different life cycles. When components process the previous data, that is to say, during the generation phase of intermediate data, the system records the generation location of intermediate data and transfers it to the next component. After the execution of the process, all the intermediate data generated by the process will no longer be used and will be deleted by the system. At the same time, the intermediate data storage space of a single process has a specified upper limit. When too much intermediate data is generated, the resource manager of the process will use the Least Recently Used algorithm (LRU) \cite{9}to clear the data, so as to prevent the overflow of memory caused by too much intermediate data. 

In order to ensure the IO efficiency of intermediate data, Alluxio\cite{10} is used as the intermediate storage reservoir to store all the intermediate data in memory. Alluxio is a virtual distributed storage system based on memory, which can greatly accelerate the speed of data reading and writing.

\subsection{Implementation Method of Business Components of Machine Learning}

\subsubsection{Machine Learning Analysis Components Based on SparkMLlib}
In section 3.2, the design of the machine learning module is described in detail. These modules complete main data processing and modeling functions in the form of components. In order to quickly provide as many algorithmic components as possible, only a small part of the processing program components are programmed according to the characteristics of the machine learning process, such as input and output components, data cleaning components, and so on, and a large number of the component functions are automatically converted to Spark jobs using Spark MLlib [11], which is Spark's own machine learning algorithm library, containing a large number of classification, regression, clustering, dimensionality reduction, and other algorithms. For example, to classify with the help of Random Forest, the Random Forest Classifier object with corresponding parameters is instantiated by the execution engine of the system according to the node information of the process. The fit method is used to fit the input data, and the corresponding Model object is generated. Then the model is serialized and saved through the intermediate data management module for subsequent prediction or verification components to use.

\subsubsection{Components of Sharing in Spark Context Execution Process}
There are two ways to run components in the process. One is to call as an independent Spark program and start the Spark Context once for each run. When the Spark program starts, it creates a context environment to determine the allocation of resources, such as how many threads and memory to be called, and then schedule tasks accordingly. The general machine learning process is composed of many components. It will take a lot of running time to start and switch the context. Another way is to share the same context for each process. The whole process can be regarded as a large Spark program. However, the execution engine of the system needs to create and manage the context for each process and release the context object to recover resources at the end of the process. 

In order to achieve context sharing, each component inherits SparkJobLife or its subclasses and implements the methods createInstance and execute. Figure 4 is the design and inheritance diagram of components classification, among which Transformers, Models, and Predictors are respectively the parent of data cleaning and data preprocessing model, learning and training model, validation and prediction model.

\begin{figure}[H]
\centering
\includegraphics[width=1.0\textwidth]{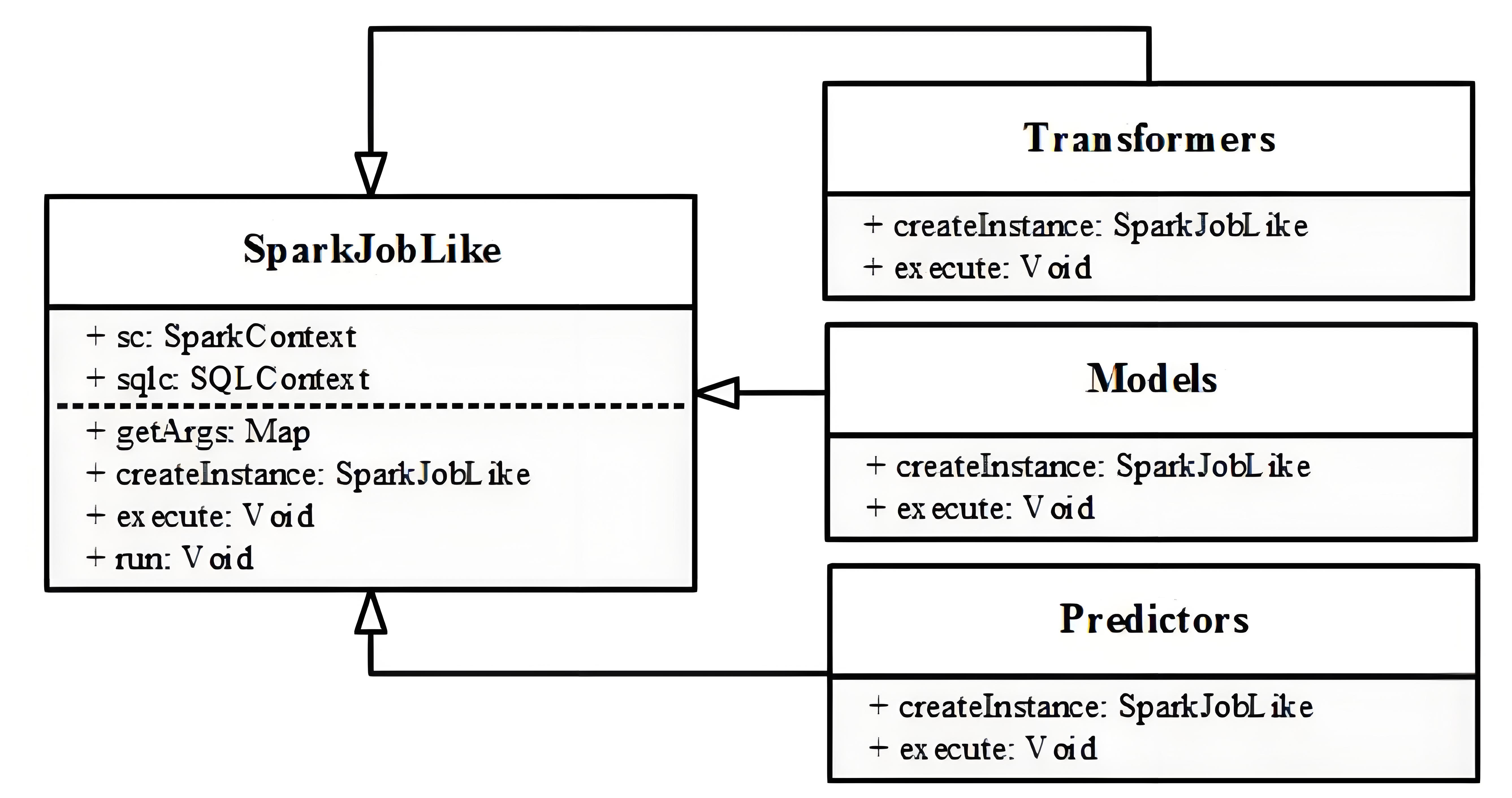}
\caption{Component class design and inheritance diagram.} \label{fig4}
\end{figure} 

\subsection{Logical Analysis Flow of Machine Learning}
\subsubsection{Process Creation}

After the user has designed and submitted the machine learning analysis process through the graphical interface, the system will start to create the logical analysis process. First, the system will make a topological analysis of the original process and generate the logical flow chart expressed by the Directed Acyclic Graph (DAG). The logical flow chart includes the dependence and parallelism of each component, as well as the input and output information and parameters.

\subsubsection{Verification Steps}
After the logical structure of the current process is generated, the validity of the overall process will be verified. The specific steps are as follows:

1. Check the input and output of each node in the graph and other necessary parameter information, and return errors if missing. For example, component users of feature processing must define input column and output column;

2. Check the integrity of the whole process to see if there is at least one input component and output component as the beginning and end, otherwise return the error;

3. Check whether there is self-circulation in the flow chart, otherwise return error;

4. Check whether each component conforms to the pre- and post-dependencies of machine learning process, for example, feature processing must be prior to model fitting, and return errors if it does not conform.

\subsection{Translation and Execution of Spark MLlib}

After validating the process, the flow chart will be submitted to the execution engine. Firstly, the system needs to represent the logical flow chart as a model that can be executed directly, and then convert it into a machine learning algorithm component based on SparkMLlib, which can be executed serially or in parallel. This process is called process translation and execution. 

MLlib \cite{11} is a distributed machine learning algorithm library with Spark built-in support, which optimizes the parallel storage and operation of large-scale data and models. With Spark MLlib, a large number of efficient component programs can be developed quickly. This section will focus on how the system translates the process into an executable model to speed up the operation of the machine learning analysis process.

\subsubsection{Translation Method for Multiple Join/fork Parallel Tasks}
Join component is a component that merges different data sets into the same data set, with a many-to-one relationship with the former component. Fork component is a component that applies the same data set to different process branches, with a one-to-many relationship with the later component. Join/fork component has a lot of applications in practice. Collaborative filtering algorithm for commodity recommendation, taken as an example, needs to join all kinds of related data such as transaction data, brand data, birth and residence information of users at the same time in order to depict user information. The specific user profile obtained is then forked to each commodity to get the corresponding preference probability \cite{12}. 

When multiple data sets join at the same time, in order to execute the process efficiently, divide-and-conquer algorithm is used to execute different join branches separately and merge them finally. When multiple process branches are generated from the same data set fork, parallel execution of each process branch will not affect the final model results. In a word, for machine learning processes with multiple join and fork tasks, it is necessary to execute in parallel as much as possible to improve operation efficiency.

\subsubsection{Translation Method of Compound Process}

The translation method used when multiple join/fork parallel tasks occur in the process was introduced in the previous section. But the actual machine learning process is not a simple serial or parallel, but a combination of serial tasks and parallel tasks, so the actual machine learning process is more complex. The difficulty to convert complex processes into execution engines lies in executing the process as parallel as possible without disrupting the data dependencies between components. The following are the translation methods for composite processes:

1. The flowchart is traversed breadth first to determine the topological relationship between business components.

2. To divide the sub-processes of the same stage according to the stages of data preprocessing, feature processing, model fitting, and prediction.

3. Critical path algorithm is used to judge the internal execution of each sub-process to determine the hierarchical relationship of branches in the sub-process.

4. The branches of the same level obtained after the last step are optimized according to the algorithm in the previous section.

\begin{figure}[H]
\centering
\includegraphics[width=1.0\textwidth]{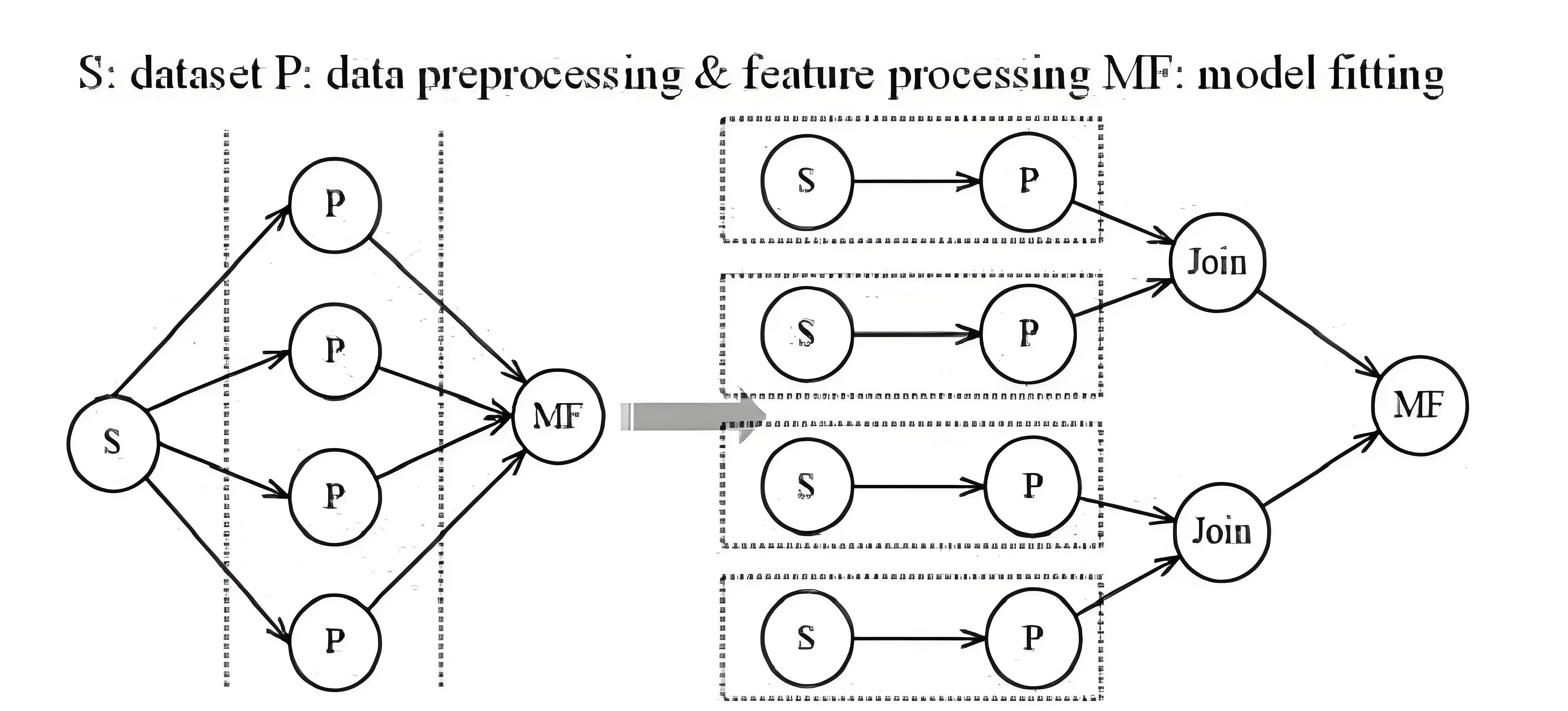}
\caption{Translation Method for Multiple Join/fork Parallel Tasks.} \label{fig4}
\end{figure}

\subsection{ Key Technologies of Spark Agent Based on LangGraph}
The implementation of Spark agents based on LangGraph involves the following key steps:

1. Within the LangChain framework, detailed interaction prompts are defined for Spark SQL, clarifying the role and functionality of the agent. The agent is designed to interact with the Spark SQL database, receive questions, construct and execute syntactically correct Spark SQL queries, analyze the results, and provide accurate answers based on these results. To improve efficiency and relevance, the query results are limited to a maximum of {top k} entries and can be sorted by relevant columns to highlight examples in the database. The agent only queries columns directly related to the question, uses specific database interaction tools, and relies on the information returned by these tools to construct the final response. Before executing the query, the agent performs a strict check of the query statement to ensure it is error-free. If problems are encountered, the agent rewrites the query statement and tries again. Additionally, the agent is strictly prohibited from executing any DML (Data Manipulation Language) statements, such as INSERT, UPDATE, DELETE, DROP, etc., to maintain the integrity of the database. For questions unrelated to the database, the agent will clearly respond with "I don't know."

2. By instantiating SparkSQLToolkit and passing the llm and toolkit as parameters to the create spark sql agent method, an instance of agent executor is constructed. This instance integrates four types of Spark SQL tools: QuerySparkSQLTool, InfoSparkSQLTool, ListSparkSQLTool  and
QueryCheckerTool, which provide the agent with the capability to interact effectively with the Spark SQL database.

3. The agent executor's invoke method is used to respond to user questions. Within the LangChain framework, the agent performs tasks through three core components: Thought (thinking), Action (action), and Observation (observation). Initially, the agent conducts an in-depth analysis and reasoning upon receiving input to determine the best course of action. Following this, the agent executes specific operations based on the results of its thinking. Then, the agent provides feedback and evaluates the outcomes of its actions, recording observations that serve as new inputs for the next round of thinking. Through the iterative cycle of these three steps, the agent can dynamically handle complex tasks and continuously optimize its behavior to achieve its goals.

\section{Experimental analysis}

\subsection{Experimentation Environment and Explanation of Experimentation Data}

At present, the system is still in the prototype stage. In order to test the function of the system, this paper uses a four-core processor, 8G memory, and a 64-bit Ubuntu system to deploy a pseudo-distributed environment for the experiment. 

The experimental data is from the public dataset of Kaggle \cite{13}. Through the crime record data of Los Angeles city from 2003 to 2015, the crime category is modeled. In order to facilitate the process description, three original features are selected in this paper, and the common machine learning analysis method is used to create the process. The data characteristics of the features and labels are shown in Table 1. In conclusion, the characters and labels are mainly character strings, which require data preprocessing to extract features and map them to numerical features.

\begin{figure}
\centering
\includegraphics[width=1.0\textwidth]{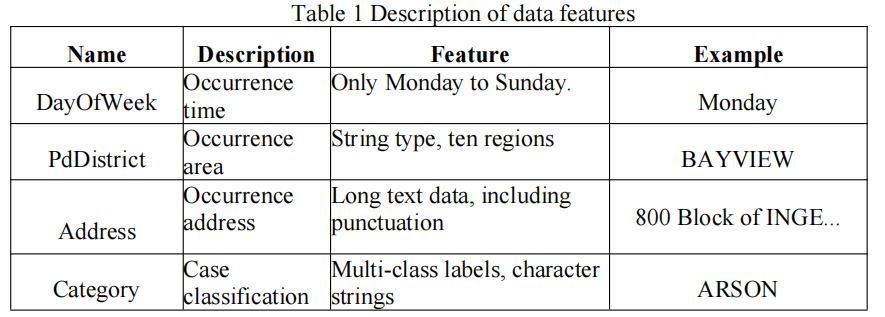}
 
\end{figure} 

\subsection{Data Preprocessing}

In order to convert the original features into numerical eigenvectors that can be computed by the training model, a series of data preprocessing tasks are needed to be implemented. In Table 2, each feature processing method is illustrated. All parameters are set by default, and any changes will be noted.

\begin{figure}[h]
\centering
\includegraphics[width=1.0\textwidth]{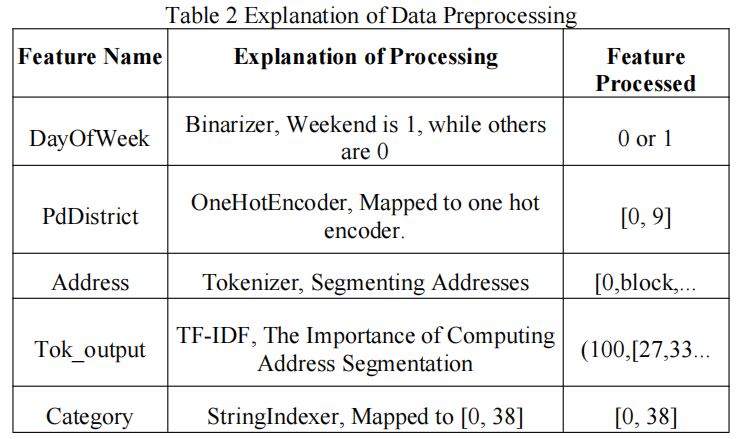}
 
\end{figure}

The feature obtained after pre-processing will be merged into feature vectors by Join components. After TF-IDF, the feature vectors have high dimensionality but are sparse. ChiSqSelector is used to select 100 features fitting models with the largest chi-square information. Logistic Regression with LBFGS is used to fit the multi-classification model, and then the test data is predicted through the trained model, and save the output results as a CSV file. The interface of the above analysis process after system creation is shown in Fig. 6.

\begin{figure}[H]
\centering
\includegraphics[width=1.0\textwidth]{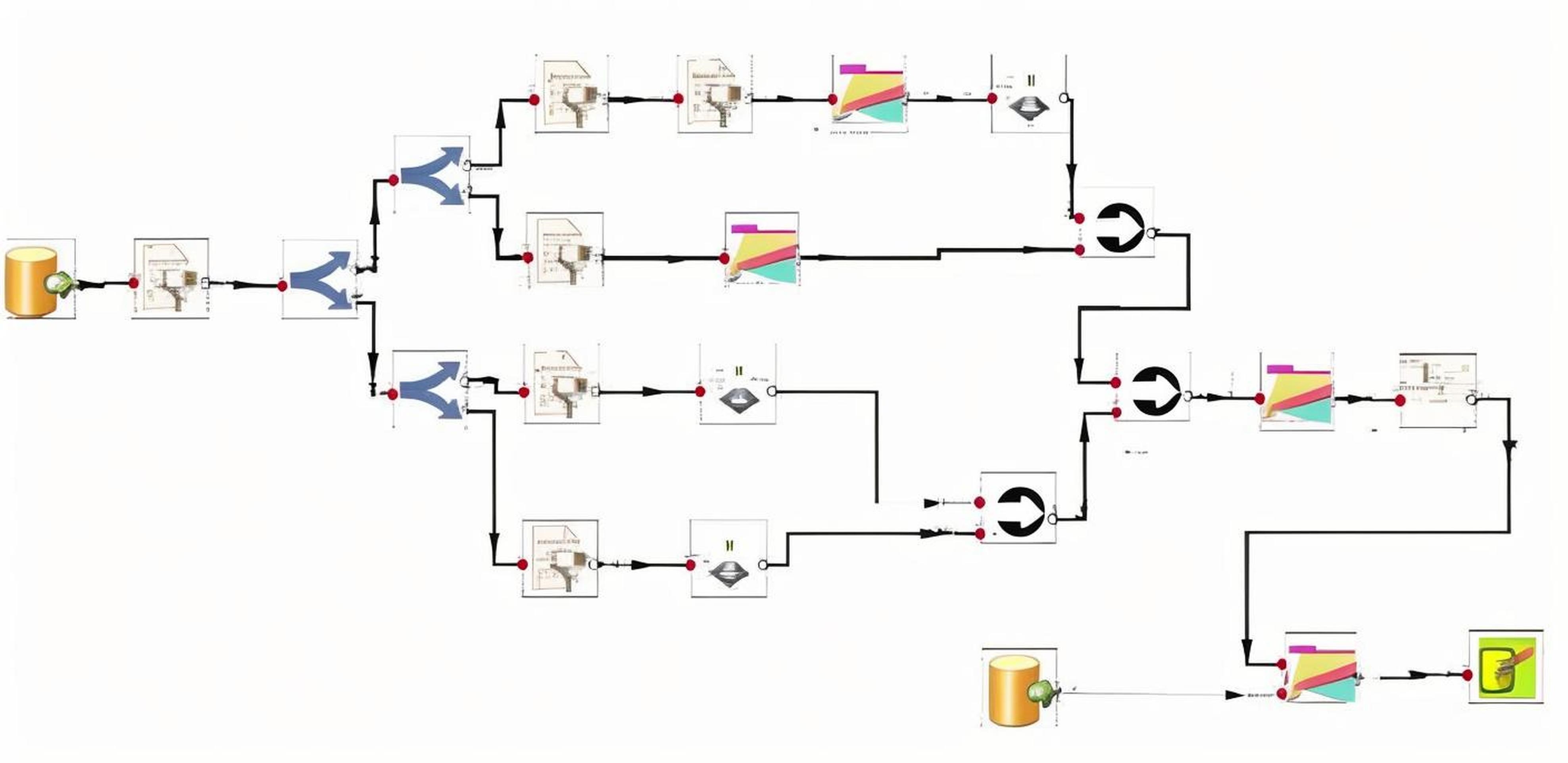}
\caption{ Created Flow Chart Interface.} \label{fig6}
\end{figure} 

\subsection{Analysis of experimental results}
By comparing the predicted value of the test data with the actual label, it is found that the accuracy is about 72.54\%. If more features are added to the process, the complexity of the model will increase, and the accuracy will also increase. With this system, machine learning processes can be created easily and quickly, and users can focus on the improvement of the analysis method. 

The parallel execution optimization of the process is introduced in Section 4. In order to test the effectiveness of the optimization method, the data of this experiment are randomly extracted and divided into ten groups of data including 10\%, 20\%, 30\%... 100\%, which are made to execute the analysis process in this experiment separately with the optimized method and the non-optimized method. No optimization refers to the sequential execution of components in the process to obtain the running time of each process in ms, as shown in Figure 7.

\begin{figure}[H]
\centering
\includegraphics[width=1.0\textwidth]{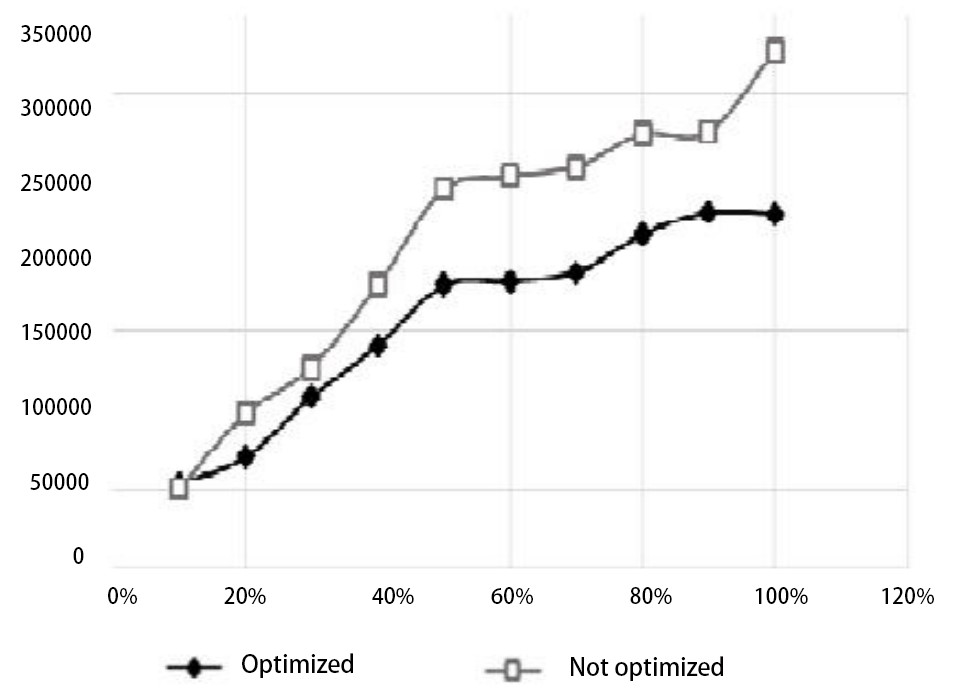}
\caption{ Comparison chart of time efficiency with optimization and with without optimization.} \label{fig7}
\end{figure} 

It can be seen that, with the linear growth of data volume, the time of non-optimized process execution increases faster, and the growth rate of time tends to increase in the later period. While with the increase of data volume, the time growth of the optimized process execution scheme is relatively slow, which shows the effectiveness of the system by implementing the optimization scheme.

\subsection{Experimental Analysis of Spark Agent Based on LangGraph}
This paper uses California housing price information as a case study to explore the practical application of Spark agents based on LangGraph. The dataset provides a wealth of information for analysis, including Longitude, Latitude, Housing Median Age, Total Rooms, Total Bedrooms, Population, Households, Median Income, and Median House Value, as shown in Figure 8.
  
 \begin{figure}
\centering
\includegraphics[width=1.0\textwidth]{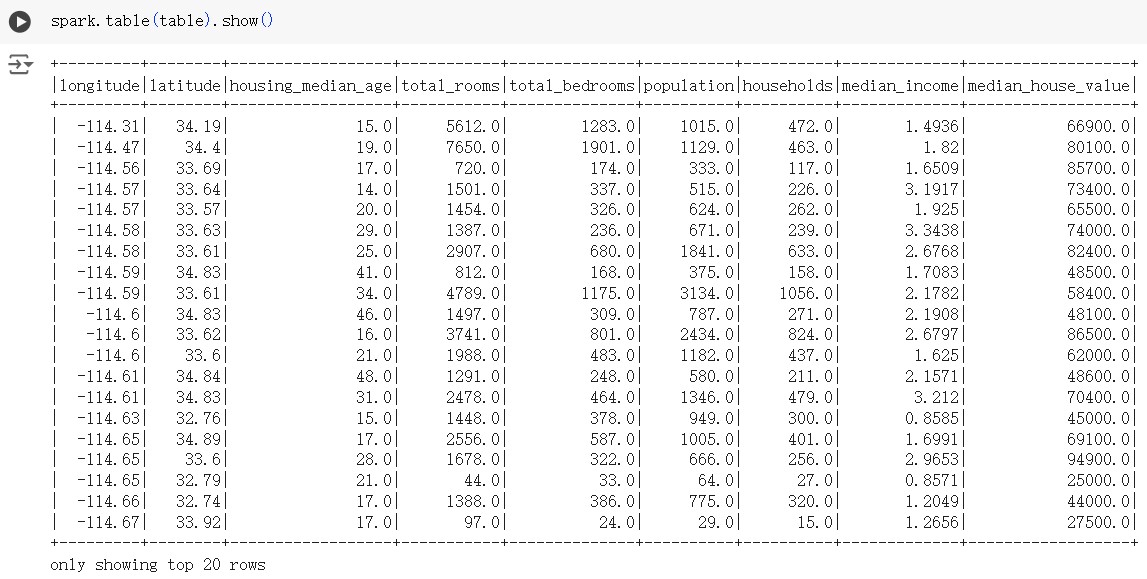}
\caption{ California Housing Price Data .} \label{fig7}
\end{figure} 

In the practical project, the California dataset was loaded via Spark, and a Spark agent was constructed using LangGraph. This agent, powered by a large-scale language model, is capable of processing and analyzing the data to provide in-depth insights into California's housing prices. The LangGraph framework utilizes a StateGraph class to define a workflow graph, which is used to construct the execution process of an agent. It allows for conditional looping between different nodes, execution of various tasks, and decision-making on whether to continue or terminate the workflow based on the response from the tools, as shown in Figure 9.
  
 \begin{figure}[H]
\centering
\includegraphics[width=0.6\textwidth]{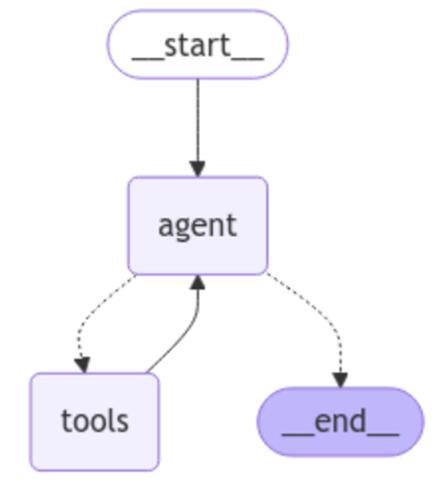}
\caption{  Agent Workflow Based on LangGraph .} \label{fig9}
\end{figure} 

As shown in Figure 10, by invoking the agentexecutor.invoke method, complex data analysis tasks within the Spark environment can be executed to retrieve information about database tables or calculate average housing prices. During the execution of the agent, the parsing of the output from the Large Language Model (LLM) plays a crucial role. If there are issues with the parsing process, it can lead to interruptions in the analysis. To ensure the accuracy of data analysis, a more powerful language model can be introduced to enhance text parsing capabilities, thereby optimizing the performance of the agent.

 \begin{figure}[H]
\centering
\includegraphics[width=1.0\textwidth]{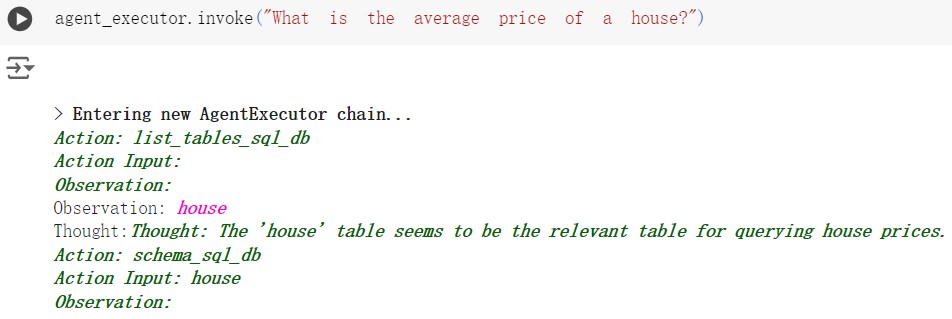}
\caption{SparkAgent Operation Example.} \label{fig10}
\end{figure}

\section{Conclusion}
In order to solve the problems that appear when data analysts use Spark to carry out machine learning analysis of large-scale data, this paper designs and implements a prototype of a distributed, flow-based Analysis System that supports multiple machine learning algorithms. In section 3 of this paper, the business model and architecture of the system was introduced as a whole. In section 4 of this paper, the key technologies from each module are described in detail, including the storage and management of intermediate data, the implementation of business components of machine learning, the creation and validation of machine learning processes, the translation and execution of machine learning processes. It also optimizes the execution of complex machine learning processes logically and translates the logical flow chart into a model that can be executed in parallel as efficiently as possible in the physical execution phase.

At present, the system converts all Spark MLlib algorithms into components automatically, which will be required to expand the algorithm library in practice. Meanwhile, in the future, relevant research can be carried out in the aspect of data dependence; for example, the system can automatically slice the data set, allocate the processing tasks of different features of the same data set to different distributed nodes for parallel processing, and improve the performance efficiency of feature processing tasks and the utilization rate of distributed resources. 

The Spark agent based on LangGraph provides a powerful tool for big data analysis systems, which not only simplifies the data analysis process but also enhances the scalability of the system. With the continuous advancement of Spark agent technology, it will play an even more critical role in the fields of data analysis and machine learning in the future.

 

 

\end{document}